\documentclass[conference]{IEEEtran}
\IEEEoverridecommandlockouts
\usepackage{cite}
\usepackage{amsmath,amssymb,amsfonts}
\usepackage{algorithmic}
\usepackage{graphicx}
\usepackage{textcomp}
\usepackage{xcolor}
\def\BibTeX{{\rm B\kern-.05em{\sc i\kern-.025em b}\kern-.08em
    T\kern-.1667em\lower.7ex\hbox{E}\kern-.125emX}}
\begin{document}

\title{QMGeo: Differentially  Private Federated Learning via Stochastic Quantization with Mixed
Truncated Geometric Distribution
}

\author{\IEEEauthorblockN{ Zixi Wang, M. Cenk Gursoy}
\IEEEauthorblockA{\textit{Department of Electrical Engineering and Computer Science, Syracuse University, Syracuse, NY}
\\
\{zwang227, mcgursoy\}@syr.edu}}

\maketitle

\begin{abstract}
    Federated learning (FL) is a framework which allows multiple users to jointly train a global machine learning (ML) model by transmitting only model updates under the coordination of a parameter server, while being able to keep their datasets local. One key motivation of such distributed frameworks is to provide privacy guarantees to the users. However, preserving the users' datasets locally is shown to be not sufficient for privacy. Several differential privacy (DP) mechanisms have been proposed to provide provable privacy guarantees by introducing randomness into the framework, and majority of these mechanisms rely on injecting additive noise. FL frameworks also face the challenge of communication efficiency, especially as machine learning models grow in complexity and size. Quantization is a commonly utilized method, reducing the communication cost by transmitting compressed representation of the underlying information. Although there have been several studies on DP and quantization in FL, the potential contribution of the quantization method alone in providing privacy guarantees has not been extensively analyzed yet. We in this paper present a novel stochastic quantization method, utilizing a mixed geometric distribution to introduce the randomness needed to provide DP, without any additive noise. We provide convergence analysis for our framework and empirically study its performance. 
\end{abstract}


\section{Introduction}
The growing prevalence of machine learning (ML) applications underscores the critical need for a vast and diverse corpus of training data. Conventional centralized ML models, while effective, necessitate the central storage and processing of data, which raises significant privacy concerns and presents formidable challenges when dealing with devices and sensors equipped with limited data transmission capabilities and resources. In response to the above-mentioned challenges, federated learning (FL) has been intensively studied recently, as an approach where machine learning (ML) models of different users are jointly trained without requiring the sharing of the raw local data \cite{kairouz2021advances}\cite{FLOrigin}. Specifically, FL supports a distributed ML framework in which each user trains its own local ML model under the coordination of a central server, that combines the local models by requiring the users to transmit only their model updates. With the users' datasets remaining local, FL avoids the heavy communication cost and privacy leakage of transmitting entire datasets. 

As mentioned above, one of the key motivations of FL structure is to reduce the communication cost. However, as modern ML applications start involving larger models, even transmitting only the model update at high precision can lay a heavy burden on communication resources, especially for edge devices whose transmission capabilities are often limited. This has led to quantization being widely used to overcome this challenge in FL. In particular, the updates from the users are first quantized into an efficient representation before the transmission\cite{FL-LossyQuant}\cite{FL-VecQuant}.

In addition to being a more communication-efficient scheme, another key motivation behind FL is that it alleviates privacy concerns to a degree by requiring only local processing of user data. However, while FL facilitates the users' data to remain decentralized at the users' local devices, this is not necessarily sufficient to ensure complete privacy. Recent studies have shown that by analyzing the model updates that get transmitted through the air \cite{PrivSurvey} or just the final model itself \cite{MembershipInfer} \cite{MembershipInfer-other}, privacy leakage can occur. Differential privacy (DP)\cite{DP-Dwork} is almost a de-facto mechanism used for privacy guarantees in  ML applications. DP provides privacy guarantees in FL systems by introducing random perturbation to the transmitted model update, thus introducing randomness and perturbation to conceal the sensitive information about user data. DP mechanism has already been implemented in many real world applications\cite{RAPPOR}\cite{Micro-DataCollect} \cite{Census}. 

We note that quantization methods lead to a certain amount of distortion in the information that the transmitted model updates were able to encode. Since DP methods ensure privacy by introducing random perturbation, the next question follows naturally: If a quantization method is stochastic, would the quantization process itself be able to provide provable DP guarantee? Our work is primarily motivated by this question.

\subsection{Related Work}
There exist studies that jointly consider reducing communication cost by quantization and providing DP guarantees. For instance, the authors in \cite{cpsgd} propose applying the binomial mechanism with quantization to achieve both communication efficiency and differential privacy, with an improved analysis showing better utility for the binomial mechanism. \cite{Descrete-Gaussian} efficiently discretizes and flattens the model updates from the users and adds discrete Gaussian noise. However, these mechanisms still rely heavily on additional noise injection to achieve differential privacy. 

The authors in \cite{Poisson-Binomial} introduce a novel mechanism where the model update information is first encoded into a parameter of a binomial distribution, and the mechanism generates samples from the distribution, without the need of additive noise. \cite{Decentralized-Quant-DP} also considers utilizing ternary quantization for privacy. However, their analysis applies to ternary quantization only and the DP guarantee they achieve is determined by the $l$-infinity norm of the model update.  

\subsection{Contributions}
In this paper, we propose a novel randomized quantization method that provides DP guarantees for multiple configurations of the quantization method without relying on any additional noise injection. The main contributions of this paper are summarized as follows: 
\begin{itemize}
  \item We propose a novel stochastic scalar quantization method QMGeo that utilizes the geometric distribution and no other additional noise to achieve $\epsilon$-DP, and we provide a R{\'e}nyi DP analysis as well.   
  \item We provide privacy analysis for the QMGeo method, when applied to scalar and vector models. 
  \item We provide the optimality gap analysis to show how the FL framework converges.  
\end{itemize}

\section{Federated Learning Protocol}
\label{sec:FL-protocol}
We first discuss an FL system with $N$ clients and a parameter server (PS). The goal of such an FL system is to minimize a loss function $F(\boldsymbol{w})$ with regard to the model parameter $\boldsymbol{w} \in \mathbb{R}^{d}$. We denote the dataset that user $m$ possesses as $\mathcal{B}_{n}$, and define $B$ as the total number of samples in the system, i.e., $B = \sum_{n=1}^{N} | \mathcal{B}_{n} |$ . 
We denote the user loss function as $F_{n} = \frac{1}{|\mathcal{B}_{n}|} \sum_{\boldsymbol{u} \in \mathcal{B}_{n}} f(\boldsymbol{w},\boldsymbol{u})$, where $\boldsymbol{u}$ corresponds to one particular sample from the dataset $\mathcal{B}_{n}$. Thus, the total loss function is given as follows:
\begin{equation}
    \label{eq:totalLossFunction}
    F(\boldsymbol{w}) = \sum_{n=1}^{N} \frac{| \mathcal{B}_{n} |}{B} F_{n}(\boldsymbol{w}).
\end{equation}
 
 An iterative approach is considered to minimize the above loss function. 
 The PS first broadcasts the global model parameter $w_t$ to all the users. 
 The users apply a uniform sampling to acquire the mini batch used for training with a sampling rate $\kappa = \frac{{b}(n)}{| \mathcal{B}_{n} |}$, while $\boldsymbol{b}_t(n)$ is the set of samples with size $b_n$ drawn from the dataset $\mathcal{B}_n$ in iteration $t$. 

 The user $n$ acquires the gradient as
 \begin{equation}
    \label{eq:local_grad}
     \boldsymbol{g}_{t}(n) = \nabla F_{n}(\boldsymbol{w}_t,\boldsymbol{b}_t(n)).
 \end{equation}

In addition, an element-wise clipping is applied to each element of the gradient $\boldsymbol{g}_{t}(n)$, where each element is clipped to be within the range of $[-W_{\text{max}}, W_{\text{max}}]$: 
\begin{equation}
    \bar{\boldsymbol{g}}_{t}(n) = clip ( \boldsymbol{g}_{t}(n)).
\end{equation}

The clipped gradient $\bar{\boldsymbol{g}}_{t}(n)$ is then quantized by a quantizer $Q(\cdot)$. The quantized vector $\Delta \boldsymbol{w}_t(n) = Q(\bar{\boldsymbol{g}}_{n}(t))$ is uploaded to the PS, which then aggregates all model updates from  $n$ users and update accordingly with a learning rate $\eta$: 
 \begin{equation}
    \label{eq:globalUpdates}
     \boldsymbol{w}_{t+1} = \boldsymbol{w}_{t} - \eta \sum_{n=1}^{N} \Delta \boldsymbol{w}_n(t).
 \end{equation}

The updated global model parameter $\boldsymbol{w}_{t+1}$ is then broadcasted to all the users before the next iteration.

\section{QMGeo: Quantization Method with Mixed Geometric Distribution} \label{sec:QGeo}
In this section we propose a stochastic scalar quantization method using a truncated mixed geometric distribution for the stochasticity. The quantization method considered is a scalar quantizer, quantizing the input scalar values, required to be limited within range $\{w | w \in [-W_{\text{max}}, W_{\text{max}}] \}$, into $R$ quantization levels. The quantizer is parameterized by the number of quantization levels $R$ and the parameter $p$ for the geometric distributions used to generate the mixed geometric distribution. 
We define the quantizer as $Q_{\text{MGeo}} (\cdot): \{w | w \in [-W_{\text{max}}, W_{\text{max}}] \} \rightarrow \{Bin(0), Bin(1), \dots , Bin(R-1)\}$. 

For each entry in the vector to be quantized, there would be a clipping threshold $W_{\text{max}}$, limiting the value of the corresponding entry $w$ to be within range $[-W_{\text{max}}, W_{\text{max}}]$. Then, for each integer $r \in [0,R)$, let a bin $Bin(r)$ represent the corresponding value for the $r$-th quantization level, such that the bins evenly span the range $[-W_{\text{max}}, W_{\text{max}}]$, where
\begin{align}
    Bin(r) =   - W_{\text{max}} + \frac{2rW_{\text{max}}}{R-1}.
\end{align}
$Q_{\text{MGeo}}(w)$ takes a scalar $w$ as input and outputs a value from the set of quantization levels $\left\{Bin(0), Bin(1), \dots, Bin(r), \dots, Bin(R-1)\right\}$ as the quantized output. The output value, i.e., the chosen quantization level $Bin(r)$, is sampled from a mixed truncated geometric distribution determined by the value of the input scalar $w$. 

In the following part of this section, we first introduce the truncated geometric distribution, and then discuss how the truncated mixed geometric distribution is determined.
\subsection{Truncated Geometric Distribution}
The truncated geometric distribution is essentially a geometric distribution but supported only between the range of two truncation points $\{x| x \in \mathbb{Z}, x\in [a, b-1]\}$, where $a$ and $b$ are the left and right truncation points. In this paper, we consider only one right truncation, and thus the distribution is supported on $\{x| x \in \mathbb{Z}, x\in [1, b-1]\}$. As in \cite{TruncatedGeoVar}, the probability mass function of a truncated geometric distribution $TGeo$ with a success probability $p$, a failure probability of $q = 1-p$ and a truncation point $b$ is
\begin{equation}
\label{eq:TruncatedGeo}
    Pr\{X_{{TGeo}} = k\} = p(1-p)^{k-1} \cdot \frac{1}{1 - (1-p)^{b-1}}.
\end{equation}
The mean and variance for the truncated geometric distribution $TGeo$ are
\begin{equation}
\label{eq:MeanTruncatedGeo}
    E(X_{TGeo}) = \frac{1 - b q^{b-1} + (b-1)q^{b}}{p (1- q^{b-1})},
\end{equation}
\begin{equation}
\label{eq:VarTruncatedGeo}
    \sigma_{TGeo}^2 = \frac{(1+q^{2b})q - q^b (1+q^2)b^2 + q^{b+1}(b^2 -1) }{(1-q)^2 (1 - q^b)^2}.
\end{equation}
\subsection{Mixed Truncated Geometric Distribution}
We define a $w$ centered mixed truncated geometric distribution as follows: 
\begin{equation}
    X_{\text{MTGeo}} = \left\{
    \begin{array}{rcl}
    Bin\left(r - (X_{{TGeo1}} - 1)\right) &  \text{w.p.}   &p_{\text{mix}} \\
    Bin\left(r + X_{{TGeo2}}\right) & \text{otherwise} \\
    \end{array}
    \right. \\
\end{equation}
where $p_{\text{mix}} = \frac{Bin(r+1) - w}{Bin(r+1) - Bin(r)}$, and  $X_{{TGeo1}}$ and $X_{{TGeo2}}$ are random variables sampled from two different truncated geometric distributions, ${TGeo1}(p)$ and ${TGeo2}(p)$. The mixture probability is determined by the distance from $w$ to $Bin(r)$ and $Bin(r+1)$ respectively, where $Bin(r)$ and $Bin(r+1)$ are the neighboring quantization levels to $w$. 

Since both the input and the output of the quantization scheme has a range of $[-W_{\text{max}}, W_{\text{max}}]$, we are forced to use truncated geometric distributions to sample the decided quantization level. Thus, $X \sim {TGeo1}(p)$ is supported only on $\{1, 2, \dots, r+1\}$ and $X \sim {TGeo2}(p)$ is supported only on $\{1, 2, \dots, R-r-1\}$. We obtain the truncated geometric distributions by normalizing the probability mass on the support back to $1$,  where the geometric distribution itself is parameterized by the success probability $p$:
\begin{equation}
\label{eq:TruncateGeo1}
    Pr\{X_{{TGeo1}} = k\} = p(1-p)^{k-1} \cdot \frac{1}{1 - (1-p)^{r+1}},
\end{equation}
\begin{equation}
\label{TruncateGeo2}
    Pr\{X_{{TGeo2}} = k\} = p(1-p)^{k-1} \cdot \frac{1}{1 - (1-p)^{R-r-1}}.
\end{equation}

\begin{figure}
    \centering
    \includegraphics[width=0.48\textwidth]{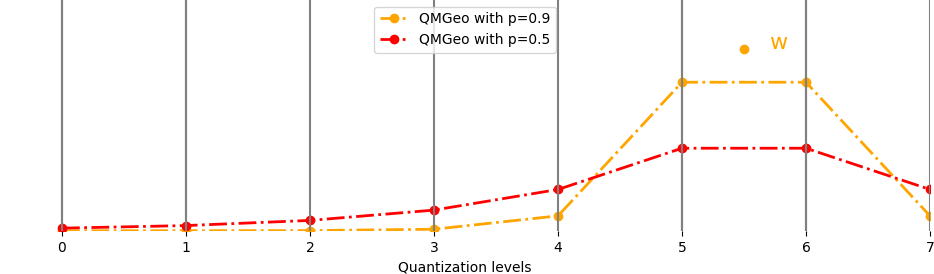}
    \caption{How $\mathcal{Q}_{\text{MGeo}}(\cdot)$ quantizes a scalar input value $w$ is demonstrated in this figure. The dotted line shows the probability of the input value $w$ being quantized to the corresponding quantization level. As shown in the figure, each quantization level is assigned with a non-zero probability. The larger $p$ is, the more skewed the distribution becomes.}
    \label{fig:quantMGeoFig}
\end{figure}

The mixed geometric distribution assigns non-zero probabilities to every quantization level, providing the randomness needed to achieve differential privacy. We control the shape of the distribution with parameter $p$. In particular, the larger $p$ is, the more skewed the distribution becomes, focusing the probability mass to neighboring quantization levels, thus less variance introduced by the $Q_{\text{MGeo}}(\cdot)$ mechanism, as demonstrated in Fig \ref{fig:quantMGeoFig}. 

\section{Differential Privacy Analysis for QMGeo} \label{sec:DP}
We first give the following definitions regarding DP \cite{DP-Dwork}. 

\textbf{$\epsilon$ - differential privacy: }
For any two adjacent datasets $D, D^{'} \in \mathcal{D}$,  and any output set of $\mathcal{S} \subset \mathcal{R}$ with domain $\mathcal{D}$ and range $\mathcal{R}$, and randomized mechanism $\mathcal{M}: \mathcal{D} \rightarrow \mathcal{R}$ that satisfies $(\epsilon, \delta)$ - differential privacy, the following property must hold:

\begin{equation} \label{eq:DP-def}
    Pr[\mathcal{M}(D) \in \mathcal{S}] \leq e^{\epsilon} Pr[\mathcal{M}(D^{'}) \in \mathcal{S}].
\end{equation}

We note that more commonly, DP mechanism use the concept of $(\epsilon, \delta)$-DP, where $\delta$ is essentially a relaxation term that allows some arbitrarily small probability for the DP mechanism to fail.


\textbf{$(\alpha, \epsilon)$-R{\'e}nyi Differential Privacy (RDP)}: For any two adjacent datasets $D, D^{'} \in \mathcal{D}$,  and any output set of $\mathcal{S} \subset \mathcal{R}$ with domain $\mathcal{D}$ and range $\mathcal{R}$, and randomized mechanism $\mathcal{M}: \mathcal{D} \rightarrow \mathcal{R}$ that satisfies $(\alpha, \epsilon)$-RDP, the following must hold:
\begin{equation} \label{eq:RDP-def}
    D_{\alpha}\left(P_{\mathcal{M}(D)} || P_{\mathcal{M}(D^{'})} \right) \leq \epsilon,
\end{equation}
where $D_{\alpha}\left(P_{\mathcal{M}(D)} || P_{\mathcal{M}(D^{'})}\right)$ is the R{\'e}nyi divergence, given by
\begin{equation} \label{RenyiDiv}
    D_{\alpha}(P || Q) = \frac{1}{\alpha -1} \log \mathbb{E}_{x\sim Q} \left[ \left( \frac{P(x)}{Q(x)} \right)^{\alpha} \right].   
\end{equation}

We first establish the  guarantee that a scalar stochastic $k$-level quantization provides as a comparison. We next establish the per-element differential privacy the scalar QMGeo method provides in Section \ref{subsec:ScalarQMGeo}. We then extend the analysis to the differential privacy guarantee when applied to a model update vector. 

\subsection{Differential privacy of the stochastic $k$-level quantization}
The stochastic $k$-level quantization is a quantization method that randomly assigns the value $w$ to neighboring quantization levels, where the probability is determined by the distance from $w$ to its neighboring quantization levels. The stochastic $k$-level quantization,   similar to the $Q_{\text{MGeo}}(\cdot)$, requires the input scalar values to be limited within some range $\{w | w \in [-W_{\text{max}}, W_{\text{max}}] \}$ and the bins evenly span the range:
\begin{align}
    Bin(r) =   - W_{\text{max}} + \frac{2rW_{\text{max}}}{R-1}.
\end{align}

The quantization is done as follows: 
\begin{equation}
    Q(w) = \left\{
    \begin{array}{rcl}
    Bin\left(r +1\right) &  \text{w.p.}   &\frac{w - Bin(r)}{Bin(r+1) - Bin(r)}\\
    Bin\left(r\right) & \text{otherwise.} \\ 
    \end{array}
    \right. \\
\end{equation}

\begin{figure}
    \centering
    \includegraphics[width = 0.48\textwidth]{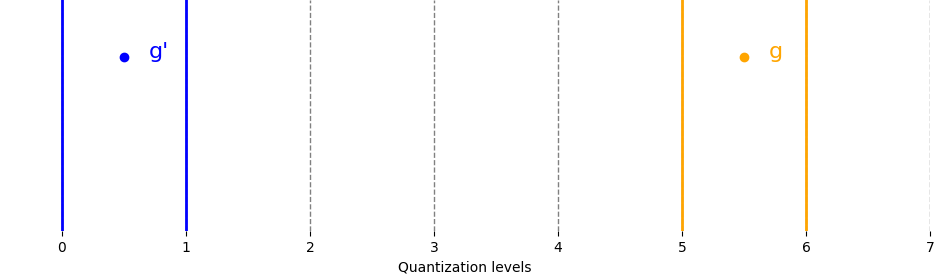}
    \caption{This figure illustrates an example of the stochastic $k$-level quantization failing to provide DP. The solid orange and blue lines are the possible output quantization levels for $g$ and $g^{'}$ respectively. It is obvious that, by observing the output of the quantization, any adversary could distinguish betwen $g$ and $g^{'}$. Any grey dotted lines are quantization levels that lies neither in the range of $Q(g)$ or $Q(g^{'})$, thus any of them being chosen as $v$ would render the $\epsilon$ unbounded.}
    \label{fig:quantFig}
\end{figure}

For number of quantization levels larger than 3, the stochastic $k$-level quantization faces a difficulty providing DP, as it could only quantize values to their neighboring quantization levels. 
When analysing the DP performance, it is obvious that if $g$ and $g^{'}$ do not lie in neighboring intervals, there will always be a choice of $v$ that achieves the worst case rendering one of the probability terms to 0 and nonzero for the other. For instance, when $v$ is a neighboring quantization level $ v = Bin(r)$ to $g$ but not for $g^{'}$, $g$ will be quantized to $v$ with a non-zero probability $p_{mix}$, while $g^{'}$ will never be quantized to $v$. In this case, $\epsilon$ cannot be bounded. An example is provided in Fig \ref{fig:quantFig}, where the intuition is that, since the quantization method quantizes to only neighboring quantization levels, there is not enough randomness to achieve DP for a number of quantization levels larger than 3:
\begin{align} \label{eq:DP-Q}
     e^{\epsilon} Pr[\mathcal{Q}(g^{'}) = v] &\geq  Pr[\mathcal{Q}(g) = v]\\
     e^{\epsilon} \times 0 &\geq p_{mix}. 
\end{align}
However, for ternary quantization, such method could provide DP guarantees to some extent \cite{Decentralized-Quant-DP}.

\subsection{Differential privacy of the scalar QMGeo method}
\label{subsec:ScalarQMGeo}
\subsubsection{$\epsilon$-DP}
We first consider the $\epsilon$-DP analysis. In our case, the considered randomized mechanism is essentially the quantization method that determines the vectors sent from the users' devices. Thus, we consider the following inequality:
\begin{equation} \label{eq:DP-QGeo}
    Pr[\mathcal{Q_{\text{MGeo}}}(g) = v] \leq e^{\epsilon} Pr[\mathcal{Q_{\text{MGeo}}}(g^{'}) = v] .
\end{equation}
(\ref{eq:DP-QGeo}) should hold for any $v$, where $g$ and $g^{'}$ are entries of gradients acquired from arbitrary neighboring datasets of a particular user. We note that the worst case scenario occurs when $v$ is either $Bin(0)$ or $Bin(R-1)$, $g$ and $g^{'}$ are equal to $Bin(0)$ and $Bin(R-1)$ respectively. Without loss of generality, we assume $v=Bin(0)$, $g=Bin(0)$ and $g^{'}=Bin(R-1)$. Substituting the values in (\ref{eq:DP-QGeo}) with the probability mass of the mixed truncated geometric distribution, we obtain
\begin{align}
\label{eq:DPDerive-QGeo}
      e^{\epsilon} &\geq  \frac{Pr[Q_{\text{MGeo}}(g) = v]}{Pr[Q_{\text{MGeo}}(g^{'})=v]}\\ 
     e^{\epsilon}  &\geq \frac{\frac{1}{2}}{\frac{1}{2}p(1-p)^{R-2} \frac{1}{1-(1-p)^{R-1}}}\\ \nonumber
    \epsilon &\geq  - \ln{p(1-p)^{R-2}} + \ln{(1-(1-p)^{R-1})}\\  \nonumber
    & = - (\ln{p} + (R-2)\ln{(1-p))} + \ln{(1-(1-p)^{R-1})}.
\end{align}

\begin{figure}
	\centering
    \includegraphics[width=0.48\textwidth]{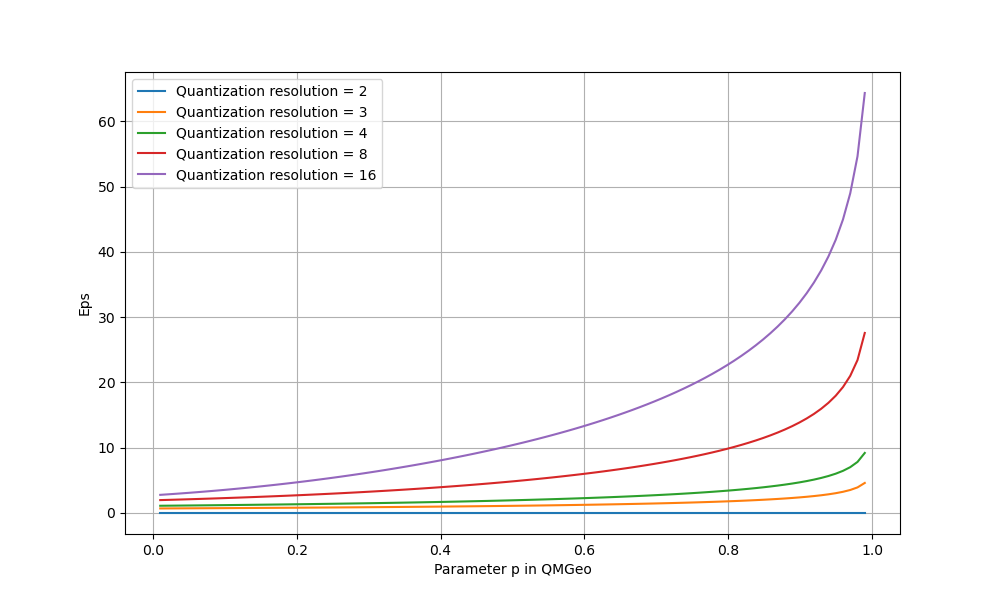}
    \caption{$\epsilon$-DP as a function of the parameter $p$ of $Q_\text{QMGeo}(\cdot)$, the success rate of the mixed truncated geometric distribution. Each curve displayed in this figure corresponds to a different number of quantization levels.} 
    \label{fig:ResCurve_Eps_vs_p}
\end{figure}

We demonstrate the trade-off between the achieved DP level $\epsilon$ and the parameters of the $QMGeo(\cdot)$ i.e., the number of quantization levels $r$ and the success probability $p$ of the mixed truncated geometric distribution. Fig \ref{fig:ResCurve_Eps_vs_p} demonstrates how the achieved DP level $\epsilon$ changes as the parameter $p$ of $Q_\text{QMGeo}(\cdot)$ changes for fixed numbers of quantization levels. $\epsilon$ increases rapidly as $p$ approaches 1, where at $p=1$, the $Q_\text{QMGeo}(\cdot)$ reduces to a conventional stochastic $r$-level quantization which could not achieve DP except for when quantizing to 3 levels or less. As shown in (\ref{eq:DPDerive-QGeo}), $\epsilon$ increases linearly with the number of quantization levels $R$. 

\subsubsection{$(\alpha, \epsilon)$-RDP}
With the quasi-convex property of the R{\'e}nyi divergence, we can show that the worst case scenario is attained by the extremal points where two input elements each sit at $Bin(0)$ and $Bin(R-1)$, where the two discrete distributions are essentially a standard truncated geometric distribution $X$ and a slightly altered one, $X^{'} = -X + R + 1$, sharing the same parameter of success rate $p$ and supported on the same set of quantization levels. We then substitute parameters $R$, $p$ from the $\mathcal{Q}_{\text{MGeo}(\cdot)}$ into (\ref{eq:RDP-def}) and obtain
\begin{align} \label{eq:RDP-derive}
    &D_{\alpha}\left(P_{\mathcal{M}(D)} || P_{\mathcal{M}(D^{'})} \right) \leq D_{\alpha}\left(P_{X} || P_{X^{'}} \right) \\
    &=\frac{1}{\alpha -1} \log \left\{ \frac{1}{2}\frac{(1-q^{R-1})^{\alpha-1}}{(q^{R}p)^{\alpha-1}} + \frac{1}{2}(pq^{R-1}\frac{1}{1-q^{R-1}})^{\alpha} \right. \\ \nonumber
    &+ \left. \frac{pq^{-2\alpha+(1-\alpha)R+1}}{2(1-q^{R-1})} \cdot \frac{q^{4\alpha-2}(1-q^{(2\alpha-1)(R-2)})}{1-\ q^{4\alpha-2}} \right\}
\end{align}
where $q=1-p$. Thus, we have
\begin{align} \label{eq:RDP-Scalar}
    \epsilon(\alpha) &\geq \frac{1}{\alpha -1} \log \left\{ \frac{1}{2}\frac{(1-q^{R-1})^{\alpha-1}}{(q^{R}p)^{\alpha-1}} + \frac{1}{2}(pq^{R-1}\frac{1}{1-q^{R-1}})^{\alpha} \right. \\ \nonumber
    &+ \left. \frac{pq^{-2\alpha+(1-\alpha)R+1}}{2(1-q^{R-1})} \cdot \frac{q^{4\alpha-2}(1-q^{(2\alpha-1)(R-2)})}{1-\ q^{4\alpha-2}} \right\}.   
\end{align}

\begin{figure}
    \centering
    \includegraphics[width=0.48\textwidth]{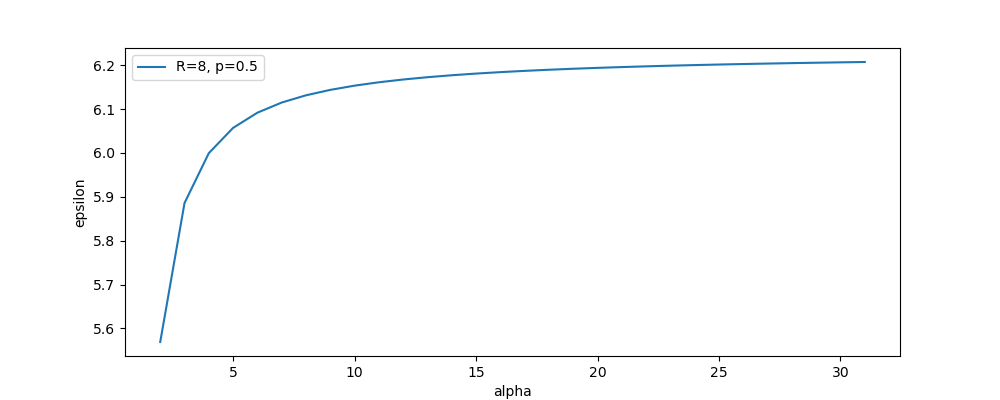}
    \caption{ Plot of $\epsilon(\alpha)$ vs. $\alpha$. The y-axis is the achieved $\epsilon$, and the x-axis is the corresponding $\alpha$. The curve is obtained with a $\mathcal{Q}_{\text{MGeo}}(\cdot)$ with $R=8$ $p=0.5$. }
    \label{fig:Fig_alphaCurve_RDPforScalar_fixedR8p5}
\end{figure}


Fig \ref{fig:Fig_alphaCurve_RDPforScalar_fixedR8p5} plots the $\epsilon(\alpha)$ curve obtained with a $\mathcal{Q}_{\text{MGeo}}(\cdot)$ with $R=8$ and $p=0.5$. 

\subsection{Differential privacy for multidimensional QMGeo}
\subsubsection{$\epsilon$-DP}
For any individual agent's gradient, we apply the scalar $Q_\text{QMGeo}(\cdot)$ to every element in the gradient. The entire privacy mechanism can be viewed as a function of multiple $Q_\text{QMGeo}(\cdot)$ mechanisms. Thus, for a gradient vector $\boldsymbol{g} = (g_1, g_2, ... , g_{d})$ the final quantized model update is $\left(Q_\text{QMGeo}(g_1), Q_\text{QMGeo}(g_2), ..., Q_\text{QMGeo}(g_{d})\right)$. 

Considering the privacy amplification theorem \cite{Priv-Amp} and the random sampling in Section \ref{sec:FL-protocol}, according to the sequential composition theorem of DP, the overall DP guarantee we obtain for the model update vector with dimension $d$ and sampling rate $\kappa$ is
\begin{equation}
\label{eq:DP-QGeo-Vector}
      \epsilon \geq  d \kappa (- (\ln{p} + (R-2)\ln{(1-p))} + \ln{(1-(1-p)^{R-1})}).
\end{equation}

\subsubsection{$(\alpha, \epsilon)$-RDP}
Similarly, RDP shows a composition of the same pattern \cite{RDP}. From \cite{RDP-subsample}, for $\alpha \leq 2$, the RDP enjoys $\epsilon_{\text{sampled}}(\alpha) = \mathcal{O}(\kappa^2\epsilon(\alpha))$, where $\kappa$ is the sampling rate. Applying to (\ref{eq:RDP-Scalar}), we have
\begin{align}
\label{eq:RDP-multiD}
    \epsilon(\alpha) \geq  &\kappa^2 d \frac{1}{\alpha-1} \times \\ \nonumber
   &\log \left\{ \frac{1}{2}\frac{(1-q^{R-1})^{\alpha-1}}{(q^{R}p)^{\alpha-1}} + \frac{1}{2}(pq^{R-1}\frac{1}{1-q^{R-1}})^{\alpha} \right. \\ \nonumber
    &+ \left. \frac{pq^{-2\alpha+(1-\alpha)R+1}}{2(1-q^{R-1})} \cdot \frac{q^{4\alpha-2}(1-q^{(2\alpha-1)(R-2)})}{1-\ q^{4\alpha-2}} \right\}. 
\end{align}

\section{Optimality Gap}
We next characterize how the perturbation introduced by the randomized quantization method impacts the convergence performance by characterizing the optimality gap between $F(\boldsymbol{w}_T)$ (the loss value achieved at global iteration $T$) and $F^*$, the optimal loss value.

We introduce two assumptions for this purpose.
\paragraph{Assumption 1}$L$-smoothness is assumed for the loss function $F$.
\paragraph{Assumption 2}(Polyak-Lojosiewicz Inequality) We assume that the loss function $F(\boldsymbol{w})$ satisfies the Polyak-Lojosiewicz (PL) condition: 
\begin{equation} \label{eq:PL}
    \frac{1}{2} \| \nabla F(\boldsymbol{w}) \|^{2} \geq \mu [ F(\boldsymbol{w}) - F^{*}].
\end{equation}

We abuse the notation of $Q_{\text{MGeo}}(\boldsymbol{g}_t)$ a bit in this section to represent the operation of applying scalar $Q_{\text{MGeo}}(\cdot)$ to every element in the vector. We define $\boldsymbol{\delta}_t = Q_{\text{MGeo}}(\boldsymbol{g}_t) - \nabla F(\boldsymbol{w}_t)$, to characterize the perturbation introduced by the randomized quantization. Following the first steps of \cite{optgapproof} and taking $\boldsymbol{\delta}_t$ into account, we have
\begin{align}
    F(\boldsymbol{w}_{t+1}) &\leq F(\boldsymbol{w}_t) + \nabla F(\boldsymbol{w}_t)^T(\boldsymbol{w}_{t+1} - \boldsymbol{w}_t) \\ \nonumber
    &+ \frac{L}{2}\| \boldsymbol{w}_{t+1} - \boldsymbol{w}_t \|^2 \\ \nonumber
   &\leq F(\boldsymbol{w}_t) - \nabla F(\boldsymbol{w}_t)^T \eta (\nabla F(\boldsymbol{w}_t) + \boldsymbol{\delta}_t) \\ \nonumber
    &+ \eta^2 \frac{L}{2}\| \nabla F(\boldsymbol{w}_t) + \boldsymbol{\delta}_t \|^2 
\end{align}

\begin{align} \label{eq:OptGapBoundIntit}
    F(\boldsymbol{w}_{t+1})  \leq  F(\boldsymbol{w}) &- \eta (1 - \frac{\eta L}{2}) \| \nabla F(\boldsymbol{w}_t)^T \|^2 \\ \nonumber
    & + \eta^2 \frac{L}{2} \| \boldsymbol{\delta}_t \|^2 \\ \nonumber
    & + \eta (-1 + \eta L) \nabla F(\boldsymbol{w}_t)^T \boldsymbol{\delta}_t \nonumber.
\end{align}
From (\ref{eq:OptGapBoundIntit}), we substract $F^*$ from both sides, which gives us
\begin{align} \label{eq:OptGapBoundMid}
     F(\boldsymbol{w}_{t+1})& - F^* \\ \nonumber
     &\leq F(\boldsymbol{w}_t) - F^* - \eta (1 - \frac{\eta L}{2}) \| \nabla F(\boldsymbol{w}_t)^T \|^2\\ \nonumber
     &+ \eta^2 \frac{L}{2} \| \boldsymbol{\delta}_t \|^2 + \eta (-1 + \eta L) \nabla F(\boldsymbol{w}_t)^T \boldsymbol{\delta}_t \nonumber .
\end{align}
We next apply the PL condition to (\ref{eq:OptGapBoundMid}), which leads to 
\begin{align} \label{eq:OptGapBoundOneStep}
    F(\boldsymbol{w}_{t+1})& - F^* \\ \nonumber
    &\leq (1-2\mu \eta (1 - \frac{\eta L}{2}) ) (F(\boldsymbol{w}_t) - F^*) + \eta^2 \frac{L}{2} \| \boldsymbol{\delta}_t \|^2 \\ \nonumber
    & +\eta(-1 + \eta L) \nabla F(\boldsymbol{w}_t)^T \boldsymbol{\delta}_t. 
\end{align}

For simplicity, let 
\begin{align}
    &X = 1-2\mu \eta (1 - \frac{\eta L}{2}) \\
    &Y = \eta^2 \frac{L}{2} \| \boldsymbol{\delta}_t \|^2 \\
    &Z = \eta(-1 + \eta L) \nabla F(\boldsymbol{w}_t)^T \boldsymbol{\delta}_t.
\end{align}

Applying (\ref{eq:OptGapBoundOneStep}) iteratively gives us
\begin{align}
\label{eq:OptGapBoundFinal}
    F(\boldsymbol{w}_{tot}) - F^* \leq X^{tot} (F(\boldsymbol{w}_{0})- F^*) + \sum_{a=1}^{tot-1} (Y+Z)X^{tot-a-1}.
\end{align}
We note that for a fixed choice of parameters of $Q_{\text{MGeo}}(\cdot)$, it is obvious that the term $\boldsymbol{\delta_t}$ is bounded in norm. Thus, (\ref{eq:OptGapBoundFinal}) indicates that with a correct choice of parameters, convergence is guaranteed for the framework.

\section{Numerical Results}
The following results are obtained on MNIST dataset (10 class handwritten digit dataset) with a total of 60000 samples. The dataset is distributed as follows: 90\% of the data are randomly split into 5 users, and 10\% of the data are reserved at the PS for evaluation. The FL framework trains a multi-layer perceptron with one hidden layer of 32 nodes. We first perform principal component analysis on the dataset and reduce the dimensionality of the input data to 100 to speed up the computation. The dimension of the model update vector is $d = 3562$. 

We set batch size to $64$ to achieve a sub-sampling rate of $\kappa = 0.005333$, learning rate $\eta$ to $0.04$ and apply a clipping threshold of $0.05$ to each element in the gradient vector. 

\begin{figure}
	\centering
    \includegraphics[width=0.48\textwidth]{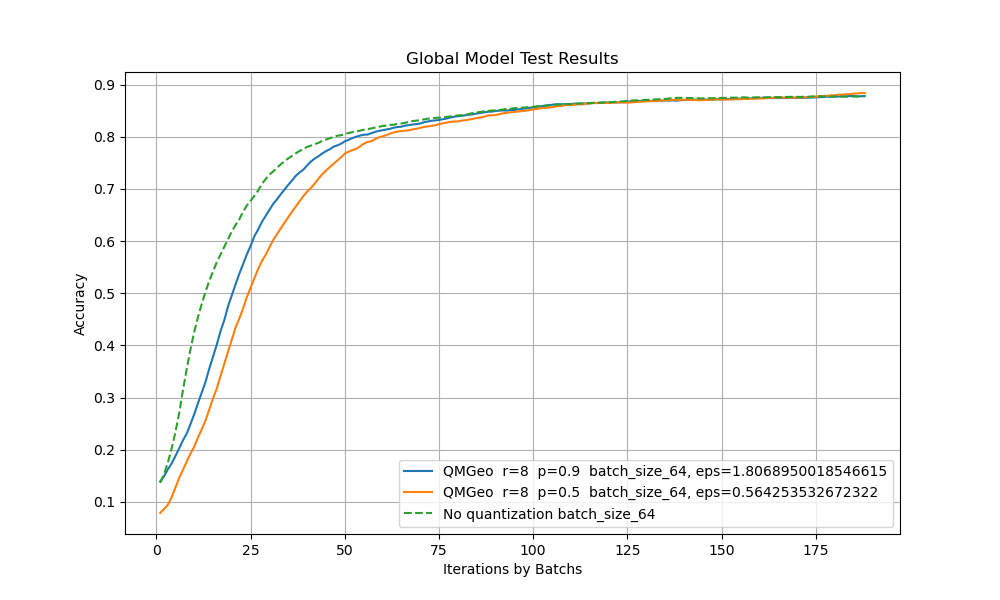}
    \caption{The two solid line curves displayed in this figure correspond to $p=0.5$ and $p=0.9$ in $Q_\text{QMGeo}(\cdot)$ respectively, with the number of quantization levels $R=8$. The dotted curve is the baseline curve where we do not apply any quantization. The y-axis shows the accuracy obtained at the global model using the hold-out test set, while the x-axis shows the number of communication rounds. The $\epsilon$ labeled in the legend is acquired using (\ref{eq:RDP-multiD}), fixing $\alpha=2$.} 
    \label{fig:B64_r8}
\end{figure}

We use (\ref{eq:RDP-multiD}) to calculate the $\epsilon(\alpha)$ per round achieved by the $Q_\text{QMGeo}(\cdot)$ method. 
Fig \ref{fig:B64_r8} demonstrates the accuracy comparison between setting $p=0.9$ and $p=0.5$, when the  number of quantization levels is fixed at $R=8$. The dotted line is a baseline case in which no quantization is performed. All curves approach a similar accuracy level, indicating that quantization does not result in performance degradation in terms of accuracy at least in the considered setup. In terms of privacy guarantees,  using $Q_\text{QMGeo}(\cdot)$ with $p=0.9$ and $R=8$ achieves $\epsilon = 1.807$ per round, and  $Q_\text{QMGeo}(\cdot)$ with $p=0.5$ and $R=8$ achieves $\epsilon = 0.564$ per round. 

\begin{figure}
	\centering
    \includegraphics[width=0.48\textwidth]{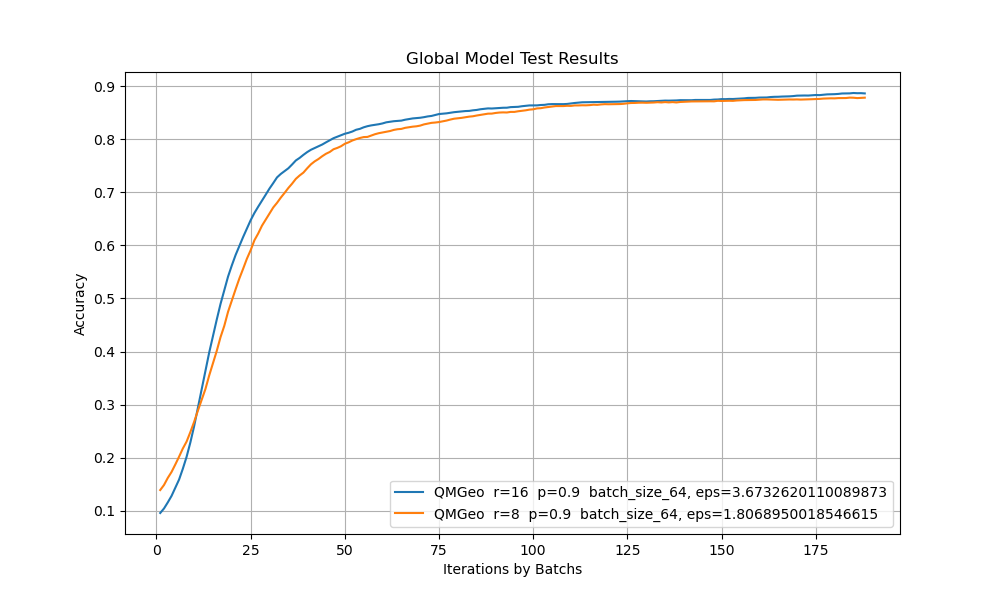}
    \caption{The two curves displayed in this figure correspond to  $R=16$ and $R=8$ in $Q_\text{QMGeo}(\cdot)$ respectively, with the parameter of success rate $p=0.9$. The y-axis shows the accuracy obtained at the global model using the hold-out test set, while the x-axis shows the number of communication rounds. The $\epsilon$ labeled in the legend is acquired using (\ref{eq:RDP-multiD}), fixing $\alpha=2$.} 
    \label{fig:B64_p9}
\end{figure}

Fig \ref{fig:B64_p9} demonstrates the accuracy comparison between setting $R=16$ and $R=8$ for fixed  parameter $p=0.9$. Using $Q_\text{QMGeo}(\cdot)$ with $p=0.9$ and $R=16$ achieves $\epsilon = 3.673$ per round, while  $Q_\text{QMGeo}(\cdot)$ with $p=0.9$ and $R=8$ leads to $\epsilon = 1.807$ per round. 

\section{Conclusions}
In this paper, we have presented a novel stochastic quantization method $Q_\text{QMGeo}(\cdot)$, that utilizes a mixture of truncated geometric distributions to provide  randomness for differential privacy. While reducing communication cost through quantization, we also achieve $\epsilon$-DP without the need of any additive noise. In particular, we have provided a privacy analysis for $Q_\text{QMGeo}(\cdot)$ both in terms of $\epsilon$-DP and RDP, and demonstrated that certain differential privacy levels can be achieved via properly designed stochastic quantization. We have further conducted an optimality gap analysis that mathematically characterizes the convergence performance of the FL framework. 


\bibliographystyle{IEEEtran}

\bibliography{refs}

\begin{thebibliography}{10}
\providecommand{\url}[1]{#1}
\csname url@samestyle\endcsname
\providecommand{\newblock}{\relax}
\providecommand{\bibinfo}[2]{#2}
\providecommand{\BIBentrySTDinterwordspacing}{\spaceskip=0pt\relax}
\providecommand{\BIBentryALTinterwordstretchfactor}{4}
\providecommand{\BIBentryALTinterwordspacing}{\spaceskip=\fontdimen2\font plus
\BIBentryALTinterwordstretchfactor\fontdimen3\font minus \fontdimen4\font\relax}
\providecommand{\BIBforeignlanguage}[2]{{%
\expandafter\ifx\csname l@#1\endcsname\relax
\typeout{** WARNING: IEEEtran.bst: No hyphenation pattern has been}%
\typeout{** loaded for the language `#1'. Using the pattern for}%
\typeout{** the default language instead.}%
\else
\language=\csname l@#1\endcsname
\fi
#2}}
\providecommand{\BIBdecl}{\relax}
\BIBdecl

\bibitem{kairouz2021advances}
P.~Kairouz and et~al., ``Advances and open problems in federated learning,'' 2021.

\bibitem{FLOrigin}
\BIBentryALTinterwordspacing
B.~McMahan, E.~Moore, D.~Ramage, S.~Hampson, and B.~A.~y. Arcas, ``{Communication-Efficient Learning of Deep Networks from Decentralized Data},'' in \emph{Proceedings of the 20th International Conference on Artificial Intelligence and Statistics}, ser. Proceedings of Machine Learning Research, A.~Singh and J.~Zhu, Eds., vol.~54.\hskip 1em plus 0.5em minus 0.4em\relax PMLR, 20--22 Apr 2017, pp. 1273--1282. [Online]. Available: \url{https://proceedings.mlr.press/v54/mcmahan17a.html}
\BIBentrySTDinterwordspacing

\bibitem{FL-LossyQuant}
\BIBentryALTinterwordspacing
M.~M. Amiri, D.~G{\"{u}}nd{\"{u}}z, S.~R. Kulkarni, and H.~V. Poor, ``Federated learning with quantized global model updates,'' \emph{CoRR}, vol. abs/2006.10672, 2020. [Online]. Available: \url{https://arxiv.org/abs/2006.10672}
\BIBentrySTDinterwordspacing

\bibitem{FL-VecQuant}
N.~Shlezinger, M.~Chen, Y.~C. Eldar, H.~V. Poor, and S.~Cui, ``{UVeQFed}: Universal vector quantization for federated learning,'' \emph{IEEE Transactions on Signal Processing}, vol.~69, pp. 500--514, 2021.

\bibitem{PrivSurvey}
\BIBentryALTinterwordspacing
M.~Nasr, R.~Shokri, and A.~Houmansadr, ``Comprehensive privacy analysis of deep learning: Passive and active white-box inference attacks against centralized and federated learning,'' in \emph{2019 {IEEE} Symposium on Security and Privacy ({SP})}.\hskip 1em plus 0.5em minus 0.4em\relax {IEEE}, may 2019. [Online]. Available: \url{https://doi.org/10.1109\%2Fsp.2019.00065}
\BIBentrySTDinterwordspacing

\bibitem{MembershipInfer}
R.~Shokri, M.~Stronati, C.~Song, and V.~Shmatikov, ``Membership inference attacks against machine learning models,'' 2017.

\bibitem{MembershipInfer-other}
\BIBentryALTinterwordspacing
S.~Truex, L.~Liu, M.~E. Gursoy, L.~Yu, and W.~Wei, ``Towards demystifying membership inference attacks,'' \emph{ArXiv}, vol. abs/1807.09173, 2018. [Online]. Available: \url{https://api.semanticscholar.org/CorpusID:50778569}
\BIBentrySTDinterwordspacing

\bibitem{DP-Dwork}
C.~Dwork, A.~Roth \emph{et~al.}, ``The algorithmic foundations of differential privacy,'' \emph{Foundations and Trends{\textregistered} in Theoretical Computer Science}, vol.~9, no. 3--4, pp. 211--407, 2014.

\bibitem{RAPPOR}
\BIBentryALTinterwordspacing
{\'{U}}.~Erlingsson, A.~Korolova, and V.~Pihur, ``{RAPPOR:} randomized aggregatable privacy-preserving ordinal response,'' \emph{CoRR}, vol. abs/1407.6981, 2014. [Online]. Available: \url{http://arxiv.org/abs/1407.6981}
\BIBentrySTDinterwordspacing

\bibitem{Micro-DataCollect}
\BIBentryALTinterwordspacing
B.~Ding, J.~Kulkarni, and S.~Yekhanin, ``Collecting telemetry data privately,'' \emph{CoRR}, vol. abs/1712.01524, 2017. [Online]. Available: \url{http://arxiv.org/abs/1712.01524}
\BIBentrySTDinterwordspacing

\bibitem{Census}
\BIBentryALTinterwordspacing
J.~Abowd, D.~Kifer, S.~L. Garfinkel, and A.~Machanavajjhala, ``Census topdown: Differentially private data, incremental schemas, and consistency with public knowledge,'' 2019. [Online]. Available: \url{https://api.semanticscholar.org/CorpusID:237407049}
\BIBentrySTDinterwordspacing

\bibitem{cpsgd}
N.~Agarwal, A.~T. Suresh, F.~Yu, S.~Kumar, and H.~B. Mcmahan, ``{cpSGD}: Communication-efficient and differentially-private distributed {SGD},'' 2018.

\bibitem{Descrete-Gaussian}
P.~Kairouz, Z.~Liu, and T.~Steinke, ``The distributed discrete gaussian mechanism for federated learning with secure aggregation,'' 2022.

\bibitem{Poisson-Binomial}
W.-N. Chen, A.~Ozgur, and P.~Kairouz, ``The poisson binomial mechanism for unbiased federated learning with secure aggregation,'' in \emph{International Conference on Machine Learning}.\hskip 1em plus 0.5em minus 0.4em\relax PMLR, 2022, pp. 3490--3506.

\bibitem{Decentralized-Quant-DP}
Y.~Wang and T.~Basar, ``Quantization enabled privacy protection in decentralized stochastic optimization,'' 2022.

\bibitem{TruncatedGeoVar}
T.~Olatayo, ``Truncated geometric bootstrap method for time series stationary process,'' \emph{Applied Mathematics}, vol. 2014, 2014.

\bibitem{Priv-Amp}
B.~Balle, G.~Barthe, and M.~Gaboardi, ``Privacy amplification by subsampling: Tight analyses via couplings and divergences,'' 2018.

\bibitem{RDP}
I.~Mironov, ``R{\'e}nyi differential privacy,'' in \emph{2017 IEEE 30th computer security foundations symposium (CSF)}.\hskip 1em plus 0.5em minus 0.4em\relax IEEE, 2017, pp. 263--275.

\bibitem{RDP-subsample}
Y.-X. Wang, B.~Balle, and S.~P. Kasiviswanathan, ``Subsampled r{\'e}nyi differential privacy and analytical moments accountant,'' in \emph{The 22nd International Conference on Artificial Intelligence and Statistics}.\hskip 1em plus 0.5em minus 0.4em\relax PMLR, 2019, pp. 1226--1235.

\bibitem{optgapproof}
S.~Ghadimi and G.~Lan, ``Stochastic first-and zeroth-order methods for nonconvex stochastic programming,'' \emph{SIAM Journal on Optimization}, vol.~23, no.~4, pp. 2341--2368, 2013.

\end{thebibliography}

\end{document}